\documentclass[letterpaper, 10 pt, conference]{ieeeconf}  

\IEEEoverridecommandlockouts                              

\usepackage{graphicx} 
\usepackage{subfigure} 

\title{\LARGE \bf
Learning effects in variable autonomy human-robot systems: how much training is enough?}

\author{Manolis Chiou$^{1}$, Mohammed Talha$^{1}$, and Rustam Stolkin$^{1}$
\thanks{$^{1}$Extreme Robotics Lab (ERL) and National Center for Nuclear Robotics (NCNR), University of Birmingham, UK}
\thanks       {\tt\small \{m.chiou, m.talha, r.stolkin,\}@bham.ac.uk}%
}

\begin{document}

\maketitle
\thispagestyle{empty}
\pagestyle{empty}

\begin{abstract}
This paper investigates learning effects and human operator training practices in variable autonomy robotic systems. These factors are known to affect performance of a human-robot system and are frequently overlooked. We present the results from an experiment inspired by a search and rescue scenario in which operators remotely controlled a mobile robot with either Human-Initiative (HI) or Mixed-Initiative (MI) control. Evidence suggests learning in terms of primary navigation task and secondary (distractor) task performance. Further evidence is provided that MI and HI performance in a pure navigation task is equal. Lastly, guidelines are proposed for experimental design and operator training practices. 
\end{abstract}
\section{Introduction}

Telerobotic systems allow humans to work in previously inaccessible areas as well as augment human capabilities in tasks requiring precision and control through teleoperation. Teleoperated robots are vital for working in unstructured and hazardous environments, in tasks such as space exploration, search and rescue (SAR), nuclear waste decommissioning and robotic surgery among others. It is commonly accepted that fully autonomous systems do not yet have the capability to deal with the unstructured nature and complex demands of these environments. Furthermore, ethical, legal, and safety reasons dictate that the "human in the loop" paradigm be followed in such applications \cite{Murphy2004,Dole2015}. Hence, teleoperated robots will see considerable use for the foreseeable future. 

However, manually teleoperated robots can be very demanding on the human operator/s both physically and mentally. For example controlling a remote robot with communications issues (e.g. time delay) and limited situational awareness (SA) to perform precise movements can be extremely difficult for operators. Additionally, teleoperation may be required while the operator is performing a secondary task simultaneously or is sleep deprived, further contributing to cognitive fatigue and errors \cite{CasperMurphy911, Murphy2005a}.

To alleviate these demands, robots can be embedded with autonomous capabilities to actively assist the operator as pointed out by several field studies \cite{CasperMurphy911, Yanco2015DARPA, Murphy2005a}. What is required is a human-robot team that benefits from the capabilities of both agents, whilst counteracting the weaknesses of each. This can be achieved with variable autonomy in which control can be traded or shared between the human operator and the robot by switching between different Levels of Autonomy (LOAs) on demand \cite{Chiou2016}. Levels of Autonomy (LOAs) refer to the degree to which the robot, or any artificial agent, takes its own decisions and acts autonomously  \cite{Sheridan1978}.

After the Fukushima Daiichi disaster, teleoperated robots were used for inspection with minimal training of participants due to the urgency of the situation \cite{guizzo_2011}. This resulted in operators having to learn skills such as opening doors, traversing rubble and stairs on-the-job under high stress and mental strain while wearing protective clothing which hampered teleoperation. For future reference, it would be useful to determine the level of training and period of time required to gain competency so training can be streamlined. It is particularly important to gain an understanding of the learning effects, training and experience of operators, as these factors are known to affect the overall robot system operating performance \cite{Armstrong2015,Bruemmer2008}. 

This paper builds upon our previous work on Human-Initiative (HI) \cite{Chiou2016} and Mixed-Initiative (MI) \cite{Chiou2019IJRR} variable autonomy control to explicitly investigate learning, training, and their correlation with performance; an area yet to be explored in the current literature. We present results from an experiment in which human operators remotely control a mobile robot conducting a search and rescue (SAR) scenario. Operators perform the task of navigation and finding a victim by using either HI or MI control while performing a cognitively demanding secondary task. A HI variable autonomy system is a system in which the human operator can dynamically switch between different LOAs based on his own judgement. This is in contrast to a MI system in which both the Robot's Artificial Intelligence (AI) and the operator have the authority to initiate LOA changes.

\section{Related work}

\textbf{Pure Teleoperation:} Drascic et al. \cite{Drascic1989} compared stereoscopic displays with monocular displays on a teleoperated Explosive Ordnance Disposal (EOD) task. They explicitly investigated learning of the task. They found that it is easier to perform highly repetitive tasks using a stereoscopic display regardless of the operator experience. Talha et al. \cite{Talha2016} presented evidence of learning effects over the course of five trials in a telemanipulation pick and place task. Due to insufficient training prior to the experiment, participants were observed to significantly improve task completion time over the trials. Additionally, they found that this led to a significantly reduced gap in performance between participants over trials. Armstrong et al. \cite{Armstrong2015} investigated how practice affects performance when driving remotely teleoperated robots through an aperture. They found that performance significantly improved with an hour of practice. They were not able to find a performance plateau (i.e. maximum performance) in their practice duration (1 hour). Contrary to our work these studies used a simple task and/or focused on pure teleoperation and didn't consider variable autonomy. 

\textbf{Shared Control:} One of the few papers providing explicit evidence of learning effects in shared control is the work of Erdogan and Argall \cite{Erdogan2017}. They showed that performance improved between sessions in a robotic wheelchair study through fewer collisions and better maneuvering. They also observed learning effects between sessions and point out the need for a longitudinal study to identify performance plateaus. Contrary to our work, the operator (i.e. the wheelchair user) is not remotely situated from the robot.

Poncela et al. \cite{Poncela2009} evaluated learning during navigation tasks using efficiency-weighted shared control. The tasks ranged from simple (e.g. wall following) to complex (e.g. passing through a doorway). 
They observed that learning was affected by task complexity. Participants performing complex tasks first, exhibited no learning later during simpler tasks whilst participants starting on simpler tasks were still learning during harder tasks but were more efficient as they could draw on knowledge from previous tasks. 

Shared control can be seen as a specific LOA inside the MI control paradigm \cite{Chiou2019IJRR}. This paper is instead focusing in dynamic LOA switching via either HI or MI. 

\textbf{Significance of Training:} Previous research has shown that for sound experiment results, it is important that participants have an equal level of competency in the system being tested through adequate training; an aspect commonly missing in other studies as pointed in \cite{Chiou2016,Talha2016}. In these studies and many others, training content and length is commonly chosen arbitrarily (e.g. \cite{Parikh2005,Carlson2010,Storms2017}) leading to learning effects during task performance and high variation in results. 

McGinn et al. \cite{McGinn2017} compared manual and semi-autonomous control in a domestic scenario involving a mobile robot navigation task. They used a relatively short training time of 10 minutes. Their statistical analysis showed that performance during manual control can vary greatly similarly to \cite{Talha2016} with participants preferring to navigate using autonomy. 

Valero-Gomez et al. \cite{Valero-Gomez2011} evaluated two adjustable autonomy techniques in an exploration task with multiple robots. They do not however provide details of the training procedure used or consider the effects of learning on performance. Similarly, most other studies (e.g. \cite{Parikh2005,Carlson2010,Gateau2016,Storms2017}) do not consider the effects of training and participant competency which has been shown to produce high variance in results \cite{Talha2016}. This is important for scientific inference as high variance can be a confounding factor affecting statistical analysis \cite{Chiou2015}.

Marble et al. \cite{Marble2004} conducted a HRI usability study of a system utilising multiple LOAs  in a SAR scenario. Their system allowed the user to switch LOA on the fly. Training consisted of 20 mins to familiarise with the system followed by execution of sample tasks until participants indicated readiness to proceed. They stated that although 20 mins is not sufficient, it is "realistic" since an operator would train once and use the system infrequently. They also found that the effect of learning meant that participants felt an improved sense of control of the robot. Interestingly, they noted that some participants rejected the autonomous modes preferring the more manual modes. The authors note this could be due to a lack of training for the respective modes and increased understanding required for their usage. Similarly Carlson and Demiris \cite{Carlson2010} provide anecdotal evidence supporting \cite{Marble2004} that poor understanding of how the robot's adaptive variable autonomy worked affected user acceptance. This further reinforces the importance of learning and training as a means to improved user acceptance of variable autonomy robotic systems.

Possibly one of the few examples of an attempt at standardized training using shared control is in \cite{Broad2017}: "This training period continues either until the user is able to successfully achieve the task three times in a row, or 15 minutes elapse." However this does not necessarily guarantee a minimum level of performance, especially the time limit. In contrast, in our work, training is enforced until all participants reach a common level of performance. However, this level is chosen arbitrarily based on pilot experiments and on our previous work. 

\subsection{Paper contribution and summary of findings}
In summary this paper is, to the best of our knowledge, the first to explicitly investigate learning effects and their relationship with training in mobile robots MI control and one of the very few (e.g. \cite{McGinn2017}) to investigate those factors in HI control. More specifically, this paper contributes: a) evidence indicating learning effects in terms of both the primary and the secondary task performance over the duration of five repeated trials; b) evidence showing that despite the five trials and the experimental duration of about 1 hour and 15 minutes, learning did not reached a plateau; c) evidence confirming results of previous work \cite{Chiou2019IJRR} that for a medium difficulty navigation task HI and MI perform equally good; d) evidence that operators prefer to use more autonomy rather than teleoperation in navigation tasks; e) insights and guidelines for informing experimental design to deal with learning effects, training and individual differences.

\section{Mixed-Initiative and Human-Initiative control}
In this paper and without loss of generality, we assume a human-robot system which has two LOAs: a) \textit{Teleoperation}, in which the human operator controls the robot manually with the joystick; b) \textit{Autonomy}, in which the operator clicks on a desired location on the map, then the robot autonomously plans and executes a trajectory to that location. The human-robot system is able to switch between these two LOAs by using two different variable autonomy methods: either Human-Initiative (HI) or Mixed-Initiative (MI) control.

The HI controller used is detailed in \cite{Chiou2016}. The controller allows the operator to switch between different LOAs on-the-fly and at any time by pressing a joystick button. It is based on the ability and authority of a human operator to initiate LOA switches based on their own judgment in order to efficiently perform a task (e.g. improving task performance or overcoming difficult performance degrading situations). The robot's AI does not have any authority to switch LOAs.

In contrast, Mixed-Initaitve (MI) control is defined in \cite{jiang2015mixed} as ``A collaboration strategy for human-robot teams where humans and robots opportunistically seize (relinquish) initiative from (to) each other as a mission is being executed, where initiative is an element of the mission that can range from low-level motion control of the robot to high-level specification of mission goals, and the initiative is mixed only when each member is authorized to intervene and seize control of it.". In the context of this paper MI control refers to the authority of both the robot's AI and the operator to initiate LOA switches.

The MI controller used was developed in our previous work \cite{Chiou2019IJRR}. This controller uses a novel expert-guided approach to initiate LOA switches. It assumes the existence of an AI task expert that given a navigational goal can provide the expected task performance. The comparison between the run-time performance of the system with the expected expert performance yields an online task effectiveness metric. This metric expresses how effective the system is performing the navigation task and it is used by the MI controller to infer if a LOA switch is needed. In practice, the expert-guided MI controller was trained to initiate LOA switches based on what human operator's did in previous HI data in order to improve performance. For further details regarding the expert-guided MI controller please refer to \cite{Chiou2019IJRR}.

\section{Experiment: evaluation of learning effects and differences in MI and HI control}

This experiment investigates within the context of HI and MI control how: a) learning (i.e. skills acquisition) affects task performance; b) learning affects LOA switching. In terms of performance we are interested in quantifying and exploring learning effects, and identifying the plateau of performance. This contributes towards identifying appropriate training procedures and give us useful insights in avoiding learning effects in future experiments, a common confounding factor. Of particular interest is the comparison of performance and learning between HI and MI as we found that our previous research \cite{Chiou2019IJRR} was possibly partly confounded with learning effects. In terms of LOA switching, we are particularly interested in how learning affects the number of LOA switches and the correlation with performance as well as the percentage of time spent in each LOA. The primary hypotheses are as follows:
 
\textbf{H1:} Performance on the primary task will improve as the number of trials increase. Participants will learn to complete the task faster as they become accustomed to the robot controls.

\textbf{H2:} Performance on the secondary task will improve as the number of trials increase. Participants will learn to complete the task faster and more accurately.

\textbf{H3:} The number of LOA switches will decrease as the number of trials increase. As participants become more comfortable with operating the robotic system, they will switch LOA more efficiency, i.e. switching only when absolutely necessary for performance. 

\subsection{Apparatus and software}

 \begin{figure}
	\centerline{\subfigure[]{\includegraphics[width=0.49\columnwidth]{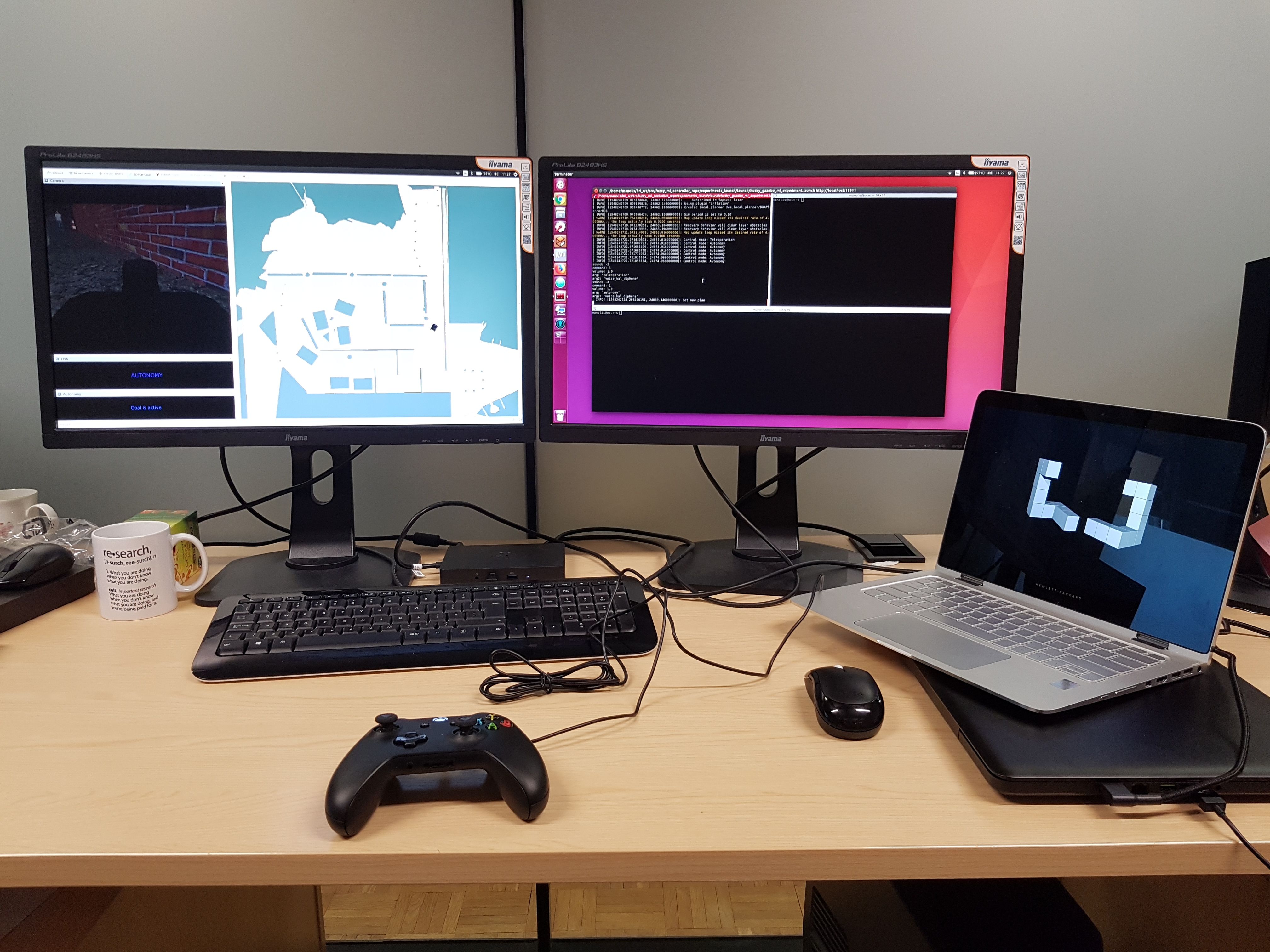}
			\label{fig:ocu}}
		\hfil
		\subfigure[]{\includegraphics[width=0.49\columnwidth]{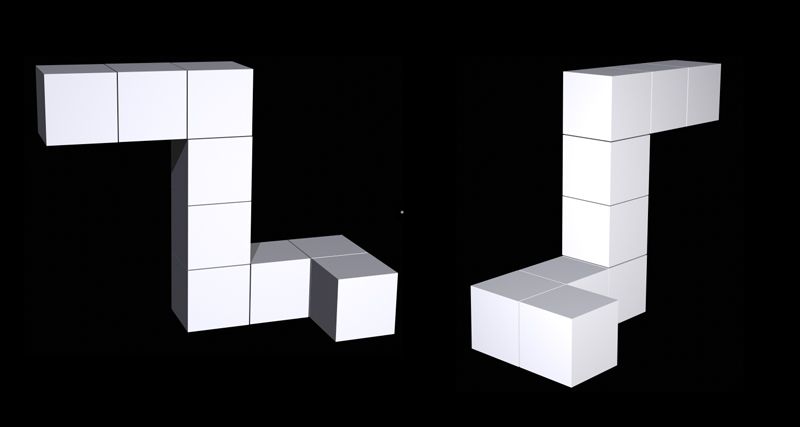}
			\label{fig:secondary_task}}}
	\caption{\textbf{\ref{fig:ocu}:} The experimental apparatus: the Operator Control Unit (OCU) composed of a laptop, a joystick, a mouse and a screen showing the GUI; and a laptop presenting the secondary task. \textbf{\ref{fig:secondary_task}:} A typical example of the secondary task.}
	\label{fig:ocu_and_secondary_image}
\end{figure}

\begin{figure}
	\centering
	\includegraphics[width=0.99\columnwidth]{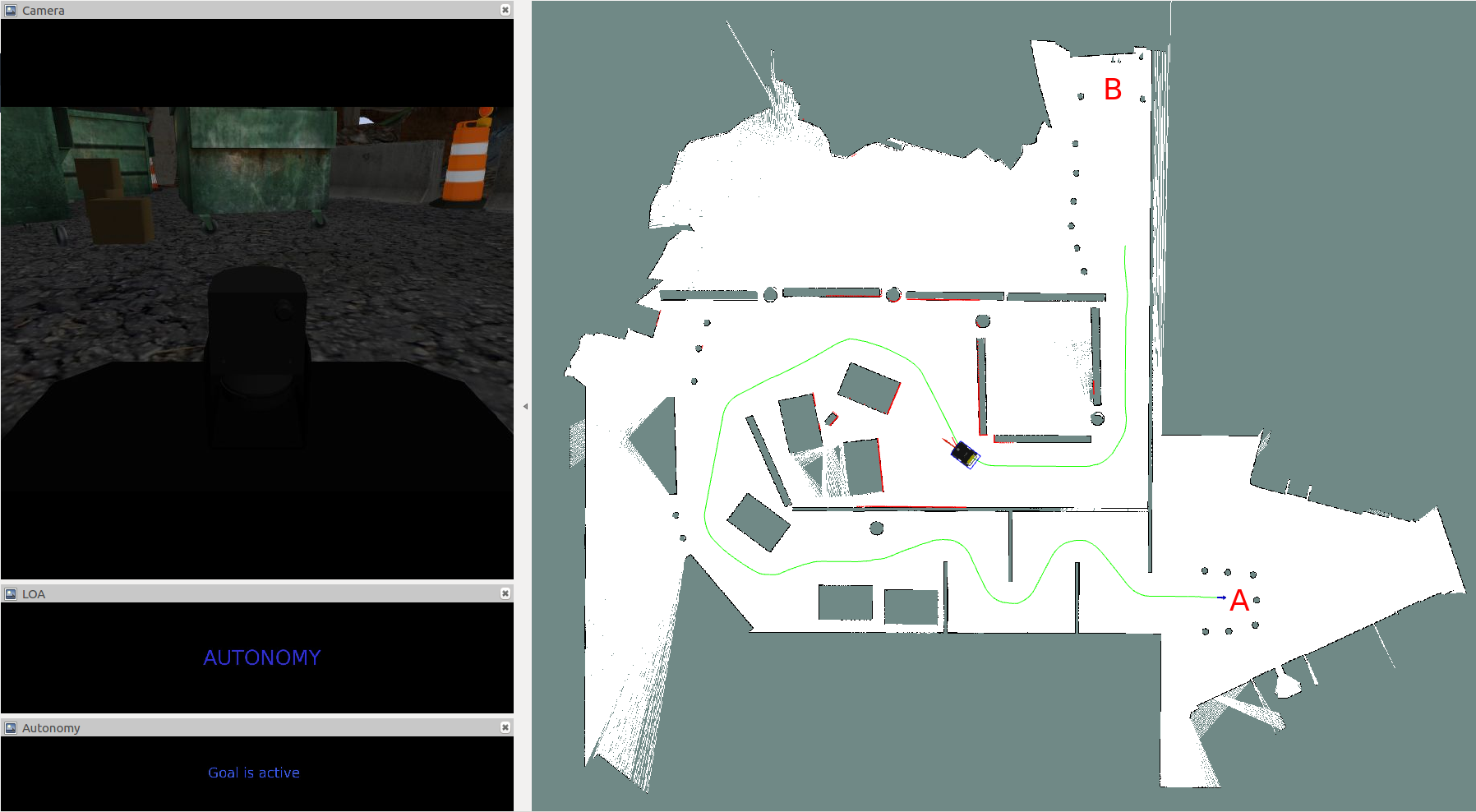}
	\caption{\textbf{Left:} video feed from the camera, the control mode in use and the status of the robot. \textbf{Right:} The map (as created by SLAM) showing the position of the robot, the current goal (blue arrow), the AI planned path (green line), the obstacles’ laser reflections (red) and the walls (black). In the map participants had to navigate from point A to point B and then back again to point A.} 
	\label{fig:gui}
\end{figure}

The environment and the robotic system used were simulated in Gazebo, a high fidelity robotic simulator. The simulated robot was a Husky equipped with a laser range finder and a RGB camera. The software used was developed in Robot Operating System (ROS) and is described in more detail in \cite{Chiou2016,Chiou2019IJRR}. The robot was controlled via an Operator Control Unit (OCU) (see Figure \ref{fig:ocu}). The OCU was composed of a mouse and a joystick as input devices, a laptop running the software and a screen showing the Graphical User Interface (GUI) (see Figure \ref{fig:gui}). 

\begin{figure}
	\centering
	\includegraphics[width=0.99\columnwidth]{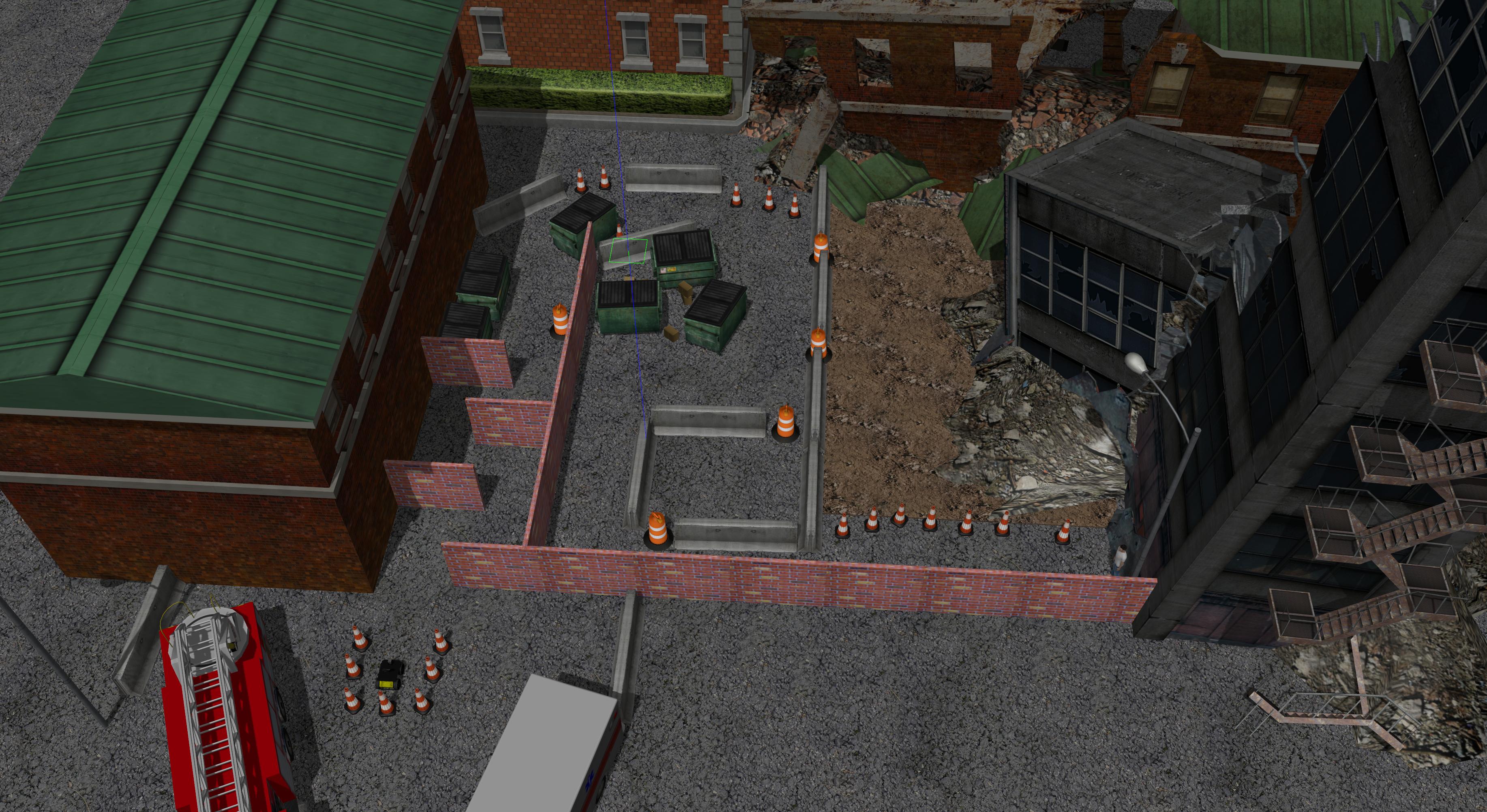}
	\caption{The test arena used, simulating a SAR scenario.} 
	\label{fig:arena}
\end{figure}

The simulation was used in order to avoid the introduction of difficult confounding factors from a real robot operating in the real world and to improve the repeatability of the experiment. This is especially true given the complexity, the length (time-wise) of the experiment, and the size of the testing arena. Examples of confounding factors include different lighting conditions affecting the video images observed by each operator and wireless communication with the robot failing due to unpredictable signal degradation (e.g. another source suddenly broadcasting in the same WiFi Chanel). The scientific merit of testing our system and investigating learning effects in a simulated environment, remains. As it can be seen in Figure \ref{fig:arena}, the simulation environment creates very realistic situations and stimuli for the participants as experienced when driving a real robot. 

The experiment's test arena was approximately $24m \times 24m$ (see Figures \ref{fig:gui} and \ref{fig:arena}) and the training arena was approximately $12m \times 6m$. Both arenas approximate a yellow coded National Institute of Standards and Technology arena \cite{nist}. 

The software used to create and display the secondary task was the OpenSesame \cite{Mathot2012OpenSesame}, a popular software used to display stimuli in psychology experiments. The images of stimuli used in the secondary task were previously created and validated for mental rotation tasks in \cite{Ganis2015}.
   
\subsection{Tasks and performance degradation factors}

The overall experiment represents a SAR scenario in which a trapped victim must be found. The experiment's participants were tasked with a primary navigation task and a cognitively demanding secondary task. In the primary task, participants had to navigate from point A in Figure \ref{fig:gui} (the beginning of the arena) to point B (the end of the arena were a victim was located) and back to point A. The path was restricted and one way, i.e. no alternative paths existed.

In the secondary task, taking place in parallel with the primary task, participants were presented with a series of images, each showing a pair of 3D objects (see Figure \ref{fig:secondary_task}). In some of the images, the objects were identical but rotated by 150 degrees. In the rest of the images the objects were mirror image objects with opposite chiralities. The operator was required to verbally state whether or not the two objects were identical (i.e. yes or no). The images were displayed in the screen of a laptop placed next to the OCU screen (see Figure \ref{fig:ocu}). 

The secondary task represents situations in which the operator needs to multitask or turn his attention elsewhere and hence robot control can be neglected. Such situations are quite common in real world robotics deployments \cite{Murphy2004}. For example robot operators must interrupt their control of the robot in order to relay information to the rescue team or to the relevant mission commander.
 
Two different kinds of performance degrading factors were used, one for each agent: artificially generated sensor noise was used to degrade the performance of autonomous navigation; and the secondary task was used to degrade the performance of the operator. In each experimental trial, each of these performance degrading situations occurred once each and at random with the restriction that they will not overlap (i.e. to appear separately from each other). This was to minimize learning effect regarding when the degrading factors appear. Thus allowing learning on task skills.

Autonomous navigation was degraded by adding Gaussian noise to the laser scanner range measurements, thereby degrading the robot’s localization and obstacle avoidance abilities. For every experimental trial this additional noise, once initiated, was lasting for 30 seconds. The secondary task was also lasting 30 seconds.

\subsection{Experimental protocol}
A total of 20 participants participated, with 10 of them controlling the robot in HI and the other 10 in MI. Before the experiment, participants completed a prior experience questionnaire assessing their experience in playing video games, operating robots, RC models and related equipment. The number of experienced participants was balanced between HI and MI.

Each operator underwent extensive standardized training before the experiment in a training arena that was different from the experiment arena. This ensured that all participants had attained a minimum skill level and understanding of the system. Firstly, the GUI and the teleoperation LOA were explained. Secondly, participants were trained to understand how distances and obstacles in the virtual environment mapped to the robot's pose, dimensions, and movement. Then, the autonomy LOA was explained. Additionally, the secondary task and the performance degrading factors were explained. Lastly, participants practiced using the MI and HI capabilities to switch between LOAs. Each system aspect was introduced gradually and participants were allowed to try them in practice. Participants were not allowed to proceed with the experimental trials until they had first demonstrated that they could complete a training obstacle course within a specific time limit, with no collisions and while presented with the two degrading factors (i.e. the secondary task and sensor noise). If participants failed in the training obstacle caorse they had to repeat it. The training lasted approximately 20 to 25 minutes depending on the participant.

After the training all participants performed the secondary task separately (i.e. without operating the robot) in order to establish baseline performance.

During the experiment each group of participants performed 5 identical trials of the same tasks (i.e. mission). 
At the end of the third trial all participants did a short 3 minutes break to prevent fatigue effects.

Participants were instructed to perform the primary task (controlling the robot in the navigation task) as quickly and safely (i.e. minimising collisions) as possible. Additionally they were instructed that when presented with the secondary task, they should do it as quickly and as accurately as possible (i.e. to have as much correct answers as possible in the 30 seconds time limit). They were explicitly told that they should give priority to the secondary task over the primary task and should only perform the primary task if the workload allowed.

The participants could only acquire SA via the GUI of the OCU which displayed the real-time video feed from the robot’s camera; the estimated robot location on the 2D SLAM map; the LOA currently active; and the status of the navigation goal (e.g. goal active, goal achieved etc).

At the end of the experiment they completed a NASA-TLX workload questionnaire \cite{Sharek2011}. NASA-TLX is an established subjective metric (i.e. questionnaire based) of perceived workload. Lastly, one of the experimenters with substantial experience and skills operating the robot, performed one trial controlling the robot via HI and one trial with MI. We will refer to him as the expert operator. His data was not included in the participant group's data but was used to allow comparison with the participants' performance. 

\section{Results}
\subsection{Statistical analysis}
Given our experimental design, when analysing learning effects in HI and in MI, data were treated as within-subject/repeated measures.
When a comparison between the HI and the MI data is taking place, these data were treated as between groups, i.e. independent samples with the type of controller as the factor with two levels: HI and MI. Given the very clear formulation of hypotheses, a series of predefined pairwise comparisons (i.e. performed explicitly on the means that were expected to differ), were performed. Taking this into account and also the small number of comparisons (one for every metric), we consider the type II errors not to be inflated. Hence, post-hoc adjustments such as Bonferroni were not used in this paper as suggested in \cite{Perneger1998}. To test the normality of the data, the Shapiro-Wilk test was used in addition to inspecting the data visually. Where data were found to conform to a normal distribution, the paired sample or independent sample t-test was used to compare the means. In cases where this did not occur, the Wilcoxon signed-rank test and the independent samples Mann-Whitney U test were used. We consider a result to be significant when it yields a $p$ value less than $0.05$. We use the term significant as referring to the statistical significance. Throughout this paper, to differentiate between the statistically significant levels, we denote with one (*) when \textit{$p < .05$}, with two (**) when \textit{$p < .01$} and with three (***) when \textit{$p < .001$}. In all graphs throughout the paper, the error bars indicate the standard error. The trial number is abbreviated as \textit{t1} for the first trial, \textit{t2} for the second and so on. The tests were two-tailed unless stated otherwise. 

\begin{figure}
	\centerline{\subfigure[]{\includegraphics[width=0.48\columnwidth]{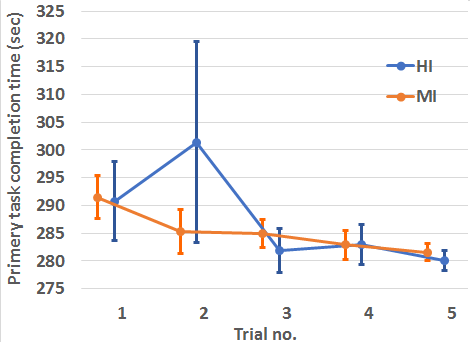}
			\label{fig:primary_task_time}}
		\hfil
		\subfigure[]{\includegraphics[width=0.49\columnwidth]{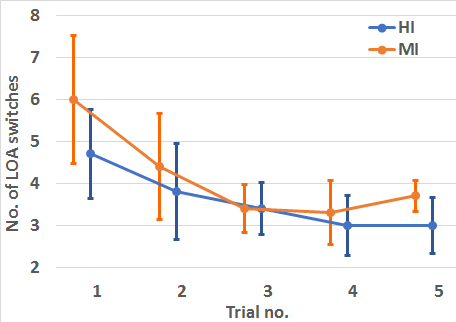}
			\label{fig:loa_switches}}}
	\caption{Primary task results: \textbf{\ref{fig:primary_task_time}:} Mean time to completion across trials for the primary task \textbf{\ref{fig:loa_switches}:} The mean number of LOA switches.}
	\label{fig:primary_task}
\end{figure}

\textbf{Primary task completion time (secs)}: Operator's in \textit{HI} completed the task faster (*) in \textit{t5} ($M = 280.1, SD = 6.06$) compared to \textit{t1} ($M = 290.8, SD = 23.8$). Similarly for \textit{MI}, operator's performance significantly (*) improved (i.e. lower primary task completion time) in \textit{t5} ($M = 281.6, SD = 5.19$) compared to the first trial \textit{t1} ($M = 291.5, SD = 12.78$). The differences in primary task time between \textit{MI} and \textit{HI} both in \textit{t1} and in \textit{t5}, are not statistically significant (see Figure \ref{fig:primary_task_time}). Analysis showed statistically significant differences between participants' performance and the expert operator in \textit{t5} both in \textit{HI} (**) and in \textit{MI} (***). The expert operator performed better than the participants with $272 sec$ in \textit{MI} and $273 sec$ in \textit{HI}.
   
\textbf{Primary task number of collisions}: In \textit{HI} the mean number of collisions in \textit{t1} was ($M = 0.5, SD = 0.7$) and in \textit{t5} was ($M = 0.1, SD = 0.32$). The expert operator had no collisions. Analyses did not show any significant difference. In \textit{MI}, \textit{t1} ($M = 0, SD = 0$) and \textit{t5} ($M = 0.1, SD = 0.32$) had no difference. The expert operator had 0 collisions, with no significant difference from \textit{t5}. Lastly, the differences in collisions between \textit{MI} and \textit{HI} both in \textit{t1} and in \textit{t5}, were not significant. All collisions took place during teleoperation.

\textbf{Number of LOA switches:} In \textit{HI}, participants did not significantly reduce the number of LOA switches as the number of trials increased (see Figure \ref{fig:loa_switches}), as no statistically significant difference was found between \textit{t1} ($M = 4.7, SD = 3.56$) and \textit{t5} ($M = 3, SD = 2.21$). In \textit{MI} the number of LOA switches between \textit{t1} ($M = 6, SD = 5.09$) and \textit{t5} ($M = 3.7, SD = 1.25$) were not statistically significant using a two-tailed Wilcoxon test. However, an one-tailed Wilcoxon test showed that \textit{t1} and \textit{t5} have significant difference (*) with participants switching LOA less frequently in \textit{t5}. Given that our hypothesis very clearly predicted the direction of the difference (i.e. lower number of switches) we consider the one-tailed Wilcoxon test to be valid. Analysis did not show any difference between \textit{MI} and \textit{HI}, both in \textit{t1} and \textit{t5}. The expert operator switched LOA 4 times during HI, which was not significantly different to the participants in \textit{t5}. Lastly, the expert operator switched LOA more (5 times) compared to participants in \textit{t5} during MI (*).

\textbf{Number of AI initiated LOA switches:} AI initiated switches are defined as LOA switches initiated by the MI controller rather than the human. Analysis did not show any difference in the mean number of AI initiated switches between \textit{t1} ($M = 1.3, SD = 1.25$) and \textit{t5} ($M = 0.6, SD = 0.52$). The number of AI initiated LOA switches for the expert operator was 0.

\begin{figure}
	\centerline{\subfigure[]{\includegraphics[width=0.49\columnwidth]{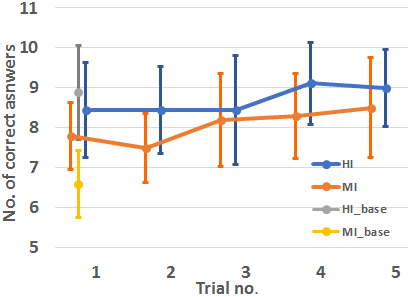}
			\label{fig:secondary_task_corr_answers}}
		\hfil
		\subfigure[]{\includegraphics[width=0.49\columnwidth]{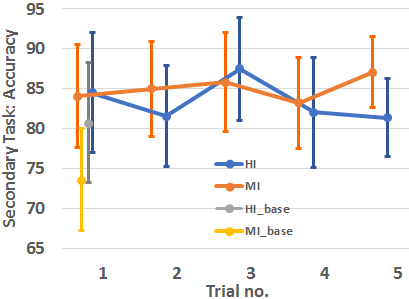}
			\label{fig:secondary_task_acc}}}
	\caption{Secondary task results: \textbf{\ref{fig:secondary_task_corr_answers}:} Mean number of correct answers across trials. \textbf{\ref{fig:secondary_task_acc}:} Mean accuracy of secondary task answers.}
	\label{fig:secondary_task}
\end{figure}

\textbf{Secondary task number of correct answers:} In \textit{HI} participants performed equally in \textit{t1} ($M = 8.8, SD = 3.42$), \textit{t5} ($M = 9.4, SD = 4.19$), and in the \textit{baseline} ($M = 9, SD = 3.56$), as no significant differences were found. In \textit{MI}, analysis showed that participants performed significantly better (**) in \textit{t5} ($M = 8.5, SD = 3.37$) compared to the \textit{baseline} ($M = 6.6, SD = 2.8$). Performance in \textit{t1} ($M = 7.8, SD = 2.86$) and \textit{t5} were similar as analysis showed no significant difference. Comparisons between \textit{HI} and \textit{MI} did not reveal significant differences in \textit{baseline}, \textit{t1}, and \textit{t5} (see Figure \ref{fig:secondary_task_corr_answers}).

\textbf{Secondary task accuracy:} In the context of this paper, the secondary task accuracy refers to the percentage of correct answers from the total responses given. In \textit{HI}, there was no significant difference between the mean accuracy in \textit{baseline} ($M = 83.68, SD = 21.04$), \textit{t1} ($M = 86.02, SD = 17.15$), and \textit{t5} ($M = 84.38, SD = 21.41$). In \textit{MI}, performance in the secondary task in terms of accuracy is significantly better (**) in \textit{t5} ($M = 87.08, SD = 13.88$) compared to the baseline ($M = 73.64, SD = 21.44$). The difference between \textit{t1} ($M = 84.05, SD = 19.85$) and \textit{t5} is not statistically significant. Comparisons between \textit{HI} and \textit{MI} did not reveal significant differences in \textit{baseline}, \textit{t1}, and \textit{t5} (see Figure \ref{fig:secondary_task_acc}).

\textbf{Percentage of time spent in autonomy LOA:} In \textit{HI}, no statistically significant difference was found in the mean percentage of time spent in autonomy between \textit{t1} ($M = 83.7, SD = 7.92$) and \textit{t5} ($M = 78.64, SD = 24.25$). In \textit{MI}, participants also spent a similar amount of percentage in autonomy in \textit{t1} ($M = 71.09, SD = 25.44$) and in \textit{t5} ($M = 77.17, SD = 21.36$). Comparisons between \textit{HI} and \textit{MI} did not reveal significant differences both in \textit{t1} and in \textit{t5}. The expert operator spent $63\%$ of the time in autonomy in \textit{HI} with no statistical difference from participants in \textit{t5}. In \textit{MI} the expert operator spent $58.8\%$ of the time in autonomy, which was significantly different (*) from participants in \textit{t5}.

\textbf{NASA-TLX and task workload:} Operators experienced less workload in \textit{HI} ($M = 31, SD = 12.74$) compared to \textit{MI} ($M = 37.53, SD = 10.52$). However, this difference was not statistical significant and thus we consider the experienced workload in \textit{HI} to be the same level as \textit{MI}.

\subsection{Discussion}
In both HI and MI control modes, participants performed better in terms of primary task completion time in trial 5 compared to trial 1. In figure \ref{fig:primary_task_time}, a trend for lower completion time can be seen as the number of trials increase in addition to a reduction in standard deviation and standard error. The above evidence suggest a learning effect (i.e. skill acquisition) for participants as they repeat the task, confirming \textbf{H1}. Additionally, the expert operator performed significantly better than the participants in trial 5. We can safely assume the expert operator's performance cannot be significantly improved in the context of this experiment and hence can be used as a performance plateau. This indicates, both in HI and in MI, that the plateau was not achieved by participants. Thus, the performance of participants could be potentially improved with more trials and/or training. 

There was no learning effect observed in HI control with respect to the mean number of LOA switches; they did not change significantly across trials. In MI, participants switched LOA significantly less in trial 5 compared to trial 1. This indicates possible learning regarding relying on the MI controller's ability to initiate LOA switches and hence actively assisting the operator. However, since the dynamics of LOA switching are complicated (e.g. participants switching LOA due to a variety of reasons \cite{Chiou2016AAI,Chiou2019IJRR}), for \textbf{H3} to be concretely confirmed further studies and evidence are required. 

The mean number of collisions both for MI and HI remained the same across control modes unaffected by learning. Analysis showed that all collisions happened during teleoperation. This is in accordance with the literature \cite{McGinn2017,Chiou2016, Chiou2019IJRR} in which the use of autonomy results in a lower number of collisions compared to pure teleoperation as the robot tends to drive more safely than humans. We consider the number of collisions observed in this experiment to be relatively low. This is likely due to the fact that participants predominately preferred to use autonomy rather than teleoperation LOA.

In HI, no learning effect was observed for the secondary task, both in terms of the number of correct answers and accuracy. Performance was on the same level across all trials. However, participants in MI performed better in trial 5 compared to the baseline both in terms of the number of correct answers and accuracy. This is potentially an important finding suggesting a learning effect in MI as hypothesised in \textbf{H2}, contrary to HI. This is likely due to participants experiencing lower cognitive loading and could rely on MI to automatically switch LOA and perform the primary task allowing the participant to focus on the secondary task.

In terms of comparing HI with MI control; performance on the primary navigation task was equal. This is important evidence further reinforcing the findings of \cite{Chiou2019IJRR}, in which the MI control performed at least as good as the HI control. Additionally, this evidence further clarifies the cases in which MI outperformed HI in \cite{Chiou2019IJRR}, this was partially due to learning effects. Perhaps a more complex task rather than simple navigation would better highlight the benefits of MI over HI. Nevertheless, there are situations in which MI control has its advantages compared to HI. For example, if the robot loses communication with the operator, the AI can take the initiative and use autonomy to return the robot to base or continue with the task. Similarly, in cases in which the operator is overwhelmed and unable to switch LOA and give commands, a robot with MI control will not wait idle and progress with the task. 

\section{Suggested guidelines for training}
This section presents insights and proposes guidelines to reduce the effects of learning, training and individual skill levels (which training is attempting to control), based on the results presented above.

Firstly, with regards to choosing an appropriate experimental design to deal with the above mentioned factors, both within-subject and between-subject designs have their merits. Within-subject designs (i.e. all participants perform all conditions) are efficient in controlling for individual differences, e.g. one participant being more competent in the task than another. For this reason such a design is preferable when the number of participants is small. On the other hand, it is more prone to learning effects and hence the use of counter balancing techniques is mandatory. Based on our experience and on the evidence presented in this paper, we propose that conditions investigated per experiment be limited to a maximum of three. This is a balanced trade-off between the number of conditions tested and learning effects and psychological effects such as boredom and fatigue becoming a confounding factor. The more conditions tested, the more likely for a difference in performance between conditions to be observed. This is due to participants continually improving in the task/system the more conditions they run or eventually getting bored.

A between-subject design can partially negate learning effects as participants perform under one condition only. It also reduces the onset of boredom and fatigue which can be introduced when carrying out multiple trials of the same repetitive task and having to fill out multiple questionnaires. However, such a design needs a large number of participants in order to deal with individual differences, which are potential extraneous factors leading to variance in results \cite{Chiou2015}. It also does not guarantee that skilled participants are distributed evenly between the treatment groups although this can be reduced by using random assignment.

Furthermore, in certain occasions it may be beneficial to use the same pool of participants. Individuals having extensive experience in operating robots in similar scenarios will require less training and their performance is less likely to vary between conditions. Hence, any performance difference when evaluating different robotics systems, will most likely be due to the specific system being tested. However, this runs the risk of reducing the scope of outcomes applying only to experienced users and not to novices.

Thirdly, with respect to the duration of learning and training; in our experiment, a performance plateau was not found despite the five repeated trials with the duration of the experiment been approximately 1 hour and 15 minutes. This is similar to  Armstrong et al. \cite{Armstrong2015} and indicates that even with substantial standardized training and prolonged exposure, participants continue to improve and hence performance measurements will keep fluctuating. Further investigation is needed but this may be due to the effects of task repetition which naturally lead to improvement in performance. Although this is expected to eventually hit a plateau, it is difficult to ascertain the exact timing as it is task and system dependant. It is not always practical to include such time consuming training to determine the plateau. It may however, depending on the task, be possible to monitor performance during training (e.g. measure the average time between subtasks) until a steady state is achieved before moving on to the live task. 

Finally, emphasis should be given to the structure of the training which must be standardized and not chosen arbitrarily. One of the reasons is that it dictates the training duration needed. Also, standardization by encoding the gradual introduction of specific concepts (e.g. control interface, robot's capabilities, and task scenario) into training steps, ensures that all participants have the minimum understanding of the task and the system.

Ultimately, training practices (e.g. structure and duration) should be informed by the specific goal of the experiment, the task, and the robotic systems used. For example, operating a 6-degree-of-freedom robotic arm is more complicated than driving a mobile robot with 2-degrees-of-freedom. Thus, training should be longer and control concepts introduced gradually. Another example is an experiment in which two robotic systems are compared. Substantial training is needed to guarantee that operators have a good understanding of the systems before using them. On the other hand, in an experiment evaluating the usability of a robotic system to novice users, perhaps it would be more meaningful to have a minimum amount of training only covering the basics of the experiment.

\section{Conclusion} 
This paper presented an experimental investigation of learning effects, training, and performance in the context of HI and MI variable autonomy control in remotely controlled mobile robots. Evidence confirms and quantifies learning effects in both HI and MI in terms of primary navigation task performance. A performance plateau was not identified for the duration of the experiment. For MI control, the observed learning effect in the secondary task in trial 5, points towards participants learning to interact with and rely on the MI controller for assistance with the primary task. This extends to LOA switching as corroborated by the reduction in the number of LOA switches. However, further investigation is needed with more participants and correlation analysis.

Additionally, evidence presented, confirmed results in previous work \cite{Chiou2019IJRR} in which equivalent experiments (i.e. similar task, difficulty, restrictions and duration) took place, that HI and expert-guided MI control perform equally good in navigation tasks. In the general case however, MI may have more benefit when the operator is overwhelmed (e.g. multiple robot to single operator scenarios) and neglects to switch to an appropriate LOA or is unable to give commands (e.g. communication failure). 

Lastly, we consider the proposed guidelines to be a key contribution of this paper for improving experimental practices regarding training and learning, and ultimately yielding results which will significantly progress our understanding of the field. As this happens, future work should move towards more complex experiments and tasks. For example, in this work, the element of surprise in relation to the secondary task and laser signal degradation (i.e. sensor noise) was missing. Participants knew that they would have to perform a secondary task or that the laser signal would become noisy, although they did not know exactly when this would happen. This was necessary in order to conduct a controlled experiment for evaluating the effects of secondary tasks on an operator.

\section*{Acknowledgment}
This work was supported by the following grants of UKRI-EPSRC: EP/P017487/1  (Remote  Sensing  in  Extreme Environments); EP/R02572X/1 (National  Centre  for  Nuclear  Robotics); EP/P01366X/1 (Robotics for  Nuclear Environments). Stolkin was also sponsored by a Royal Society Industry Fellowship.

\bibliography{IEEEabrv,ref}
\bibliographystyle{IEEEtran}

\end{document}